\documentclass{article}
\usepackage{ijcai25}

\usepackage{times}
\usepackage{soul}
\usepackage{url}
\usepackage[hidelinks]{hyperref}
\usepackage[utf8]{inputenc}
\usepackage[small]{caption}
\usepackage{graphicx}
\usepackage{amsmath,amssymb,amsfonts}
\usepackage{amsthm}
\usepackage{booktabs}
\usepackage{algorithm}
\usepackage{algorithmic}
\usepackage{xcolor}
\usepackage{tabulary}
\usepackage{multirow}
\usepackage{subfig}
\usepackage{pifont}
\usepackage[switch]{lineno}

\usepackage{multirow}
\usepackage[normalem]{ulem}
\useunder{\uline}{\ul}{}

\title{ImputeINR: Time Series Imputation via Implicit Neural Representations for Disease Diagnosis with Missing Data}

\author{
Mengxuan Li$^{1,2}$
\and
Ke Liu$^{1}$\and
Jialong Guo$^{1}$\and \\
Jiajun Bu$^{1}$\and
Hongwei Wang$^{2,}$\thanks{Corresponding authors.} \and
Haishuai Wang$^{1,*}$\\
\affiliations
$^1$Zhejiang Key Laboratory of Accessible Perception and Intelligent Systems, \\College of Computer Science,	Zhejiang University, China\\
$^2$
The Zhejiang University-University of Illinois Urbana-Champaign Institute, Zhejiang University\\
\emails
\{mengxuanli, hongweiwang\}@intl.zju.edu.cn,
\{keliu99, jialongguo, bjj, haishuai.wang\}@zju.edu.cn
}

\begin{document}

\maketitle

\begin{abstract}
Healthcare data frequently contain a substantial proportion of missing values, necessitating effective time series imputation to support downstream disease diagnosis tasks. However, existing imputation methods focus on discrete data points and are unable to effectively model sparse data, resulting in particularly poor performance for imputing substantial missing values. In this paper, we propose a novel approach, ImputeINR, for time series imputation by employing implicit neural representations (INR) to learn continuous functions for time series. ImputeINR leverages the merits of INR in that the continuous functions are not coupled to sampling frequency and have infinite sampling frequency, allowing ImputeINR to generate fine-grained imputations even on extremely sparse observed values. Extensive experiments conducted on eight datasets with five ratios of masked values show the superior imputation performance of ImputeINR, especially for high missing ratios in time series data. Furthermore, we validate that applying ImputeINR to impute missing values in healthcare data enhances the performance of downstream disease diagnosis tasks. Codes are available \textcolor{blue}{\href{https://github.com/Leanna97/ImputeINR}{here}}.
\end{abstract}

\section{Introduction}
Healthcare data inherently exhibit a temporal structure, as they are often collected sequentially over time in the form of physiological signals, electronic health records, and clinical monitoring data, making time series analysis a fundamental approach for understanding and predicting health-related outcomes \cite{schaffer2021interrupted-healthcare,wang2018learning,li2023order,li2023sccam-industrial2}. However, real-world healthcare data are frequently compromised by a substantial amount of missing values, which arise from various sources, such as sensor malfunctions, transmission errors, or irregular reporting intervals. The presence of missing values poses a significant challenge to disease diagnosis, as incomplete or corrupted data can distort model training and lead to unreliable predictions. Consequently, the imputation of missing data becomes a critical step in disease diagnosis. By filling in the gaps, imputation enables the restoration of a complete dataset, allowing diagnostic models to perform without the biases or inaccuracies that would otherwise result from missing information.

\begin{figure}[t]
\centering
\includegraphics[width = 1\linewidth]{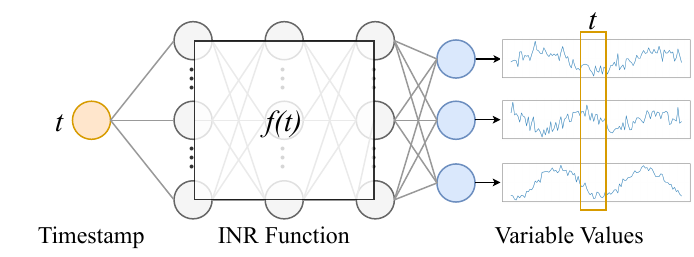}
\caption{The illustration of INR applied to time series data.}
\vspace{-1em}
\label{fig_inrts}
\end{figure}

However, most existing imputation methods do not highlight the cases of extremely sparse observed values. These works assume that the proportion of masked values requiring imputation does not exceed 50\%, which means that these methods still require a certain amount of known information. But in real-world healthcare data, the proportion of missing values is likely to be even higher. For example, the MIMIC-III dataset \cite{johnson2018mimic}, a public health-related database, contains 63.15\% missing values. The PhysioNet 2012 and 2019 challenges \cite{silva2012physionet-phy2012,reyna2019physionet-phy2019}, which involved intensive care unit (ICU) patient data, reported a missing data rate of 79.67\%, highlighting the prevalence of high missing rates in real-world healthcare datasets. Although some recent studies try to address imputation tasks in sparse scenarios \cite{alcaraz2022diffusion-sssd,tashiro2021csdi}, most work relies on diffusion-based methods, which involve the simulation of high-dimensional stochastic processes and multi-step iterative procedures. Consequently, these approaches are typically time-consuming and require a large number of training iterations, and their limited effectiveness in achieving accurate imputation can significantly impair the performance of downstream disease diagnosis tasks. How to perform imputation in extremely sparse scenarios remains challenging.

Recently, implicit neural representation (INR) has emerged as an effective method for continuously encoding diverse signals \cite{liu2023implicit-icme,molaei2023implicit}. As shown in Figure \ref{fig_inrts}, it learns continuous functions from discrete data points, mapping coordinates to signal values. By representing complex structures in a compact form, INR is not coupled to sampling frequency anymore, which allows for multi-sampling frequency inputs enabling effective feature extraction even with absent observed samples. Additionally, as a continuous function, INR has infinite sampling frequency, which means it can be queried at any coordinate. This capacity for infinite sampling frequency interpolation sets it apart from other imputation methods, making it a promising approach for fine-grained imputation. 


In this paper, we propose a novel time series imputation approach, named ImputeINR, which achieves effective imputation even in extremely sparse scenarios and the imputed data can further enhance the performance of downstream disease diagnosis tasks. Generally, we learn an INR continuous function for the target time series data and enable fine-grained interpolation with infinite sampling frequency. The parameters of the INR function are predicted by a transformer-based feed-forward network conditioned with sparsely observed values. More specifically, an adaptive group-based form of the INR function is proposed for capturing complex temporal patterns and cross-channel correlation features. It is a multilayer perceptron (MLP) composed of global layers and group layers. The former focuses on correlation information across all channels, while the latter emphasizes correlation information among variables within a single group. We observe that INR exhibits the strongest representational capacity when partitioning variables with similar distributions into the same group. Therefore, variable clustering is applied to determine the variable partition. Additionally, a multi-scale feature extraction module is incorporated to capture patterns at various temporal scales, achieving better fine-grained imputation. Experimentally, ImputeINR achieves state-of-the-art performance on eight imputation benchmarks with various ratios of masked values. Moreover, we demonstrate that the imputed data with ImputeINR, compared to other methods, significantly improves the performance of downstream disease diagnosis tasks. The major contributions of this paper are summarized as follows:
\begin{itemize}
    \item We propose ImputeINR, which learns INR continuous function to represent the continuous time series data. It achieves effective imputation in extremely sparse scenarios and the imputed data can improve the performance of downstream disease diagnosis tasks.
    \item We design an adaptive group-based form of the INR continuous function to effectively capture intricate temporal patterns and cross-channel correlation features.
    \item We apply variable clustering to determine the variable partition, allowing our group-based architecture to learn correlation information across all variables and among variables with similar distributions.
    \item Extensive experiments show that ImputeINR outperforms other baselines on eight datasets under five ratios of masked values.  We also demonstrate that using ImputeINR for healthcare data imputation significantly enhances disease diagnosis performance.
\end{itemize}

\section{Related Work}
\subsection{Time Series Imputation}
The earliest time series imputation methods are based on the statistical properties of the data, using mean/median values or statistical models to fill in missing values, such as SimpleMean/SimpleMedian \cite{fung2006methods} and ARIMA \cite{afrifa2020missing}. Then, machine learning methods learn data patterns, showing greater adaptability and accuracy. Prominent works of these approaches include KNNI \cite{altman1992introduction-knn} and MICE \cite{van2011mice}. 
Although these methods are easy to interpret, their limitations lie in capturing the complex temporal and variable information inherent in time series data. Recently, there has been widespread interest in using deep models to capture complex temporal patterns for imputation of missing values, due to their powerful representation capabilities. Common architectures include RNN-based methods (e.g., M-RNN \cite{yoon2018estimating-mrnn}, NRTSI \cite{shan2023nrtsi}, and BRITS \cite{cao2018brits}) , CNN-based methods (e.g., TimesNet \cite{wutimesnet}), MLP-based methods (e.g., DLinear \cite{zeng2023transformers-dlinear}, TimeMixer \cite{wangtimemixer}), transformer-based methods (e.g., SAITS \cite{du2023saits}, FPT \cite{zhou2023one-fpt}, iTransformer \cite{liuitransformer}, ImputeFormer \cite{nie2024imputeformer}), diffusion-based methods (e.g., CSDI \cite{tashiro2021csdi}, SSSD \cite{alcaraz2022diffusion-sssd}).

\subsection{Implicit Neural Representations}
INR uses neural networks to model signals as continuous functions rather than explicitly representing them as discrete points. It captures complex high-dimensional patterns in data by learning a continuous mapping from input coordinates to output values. Various scenarios have seen successful applications, such as 2D image generation \cite{saragadam2022miner,liu2023partition,zhang2024attention-anr}, 3D scene reconstruction \cite{yin2022coordinates-3dinr3,liu2023implicit-icme,yang2024pm-inr3d}, and video representations \cite{mai2022motion-videoinr,guo2025metanerv}. Since INR learns a continuous function, it is not coupled to the resolution, which implies that the memory needed to parameterize the signal does not depend on spatial resolution but rather increases with the complexity of the underlying signal. Also, INR has infinite resolution, which means it can be sampled at an arbitrary sampling frequency. Therefore, we leverage this characteristic of INR to perform time series imputation tasks. Sampling from the continuous function of INR enables fine-grained imputation even with extremely sparse observed data. To learn the INR for target signal, there are mainly two typical strategies: gradient-based meta-learning methods \cite{lee2021metainr,liu2023partition} and feed-forward hyper-network prediction methods \cite{chen2022transformers-transinr,zhang2024attention-anr}. In this work, we use a transformer-based feed-forward method to predict the INR for time series data since it can be easily adopted to an end-to-end imputation framework.

\begin{figure*}[t]
\centering
\includegraphics[width = \textwidth]{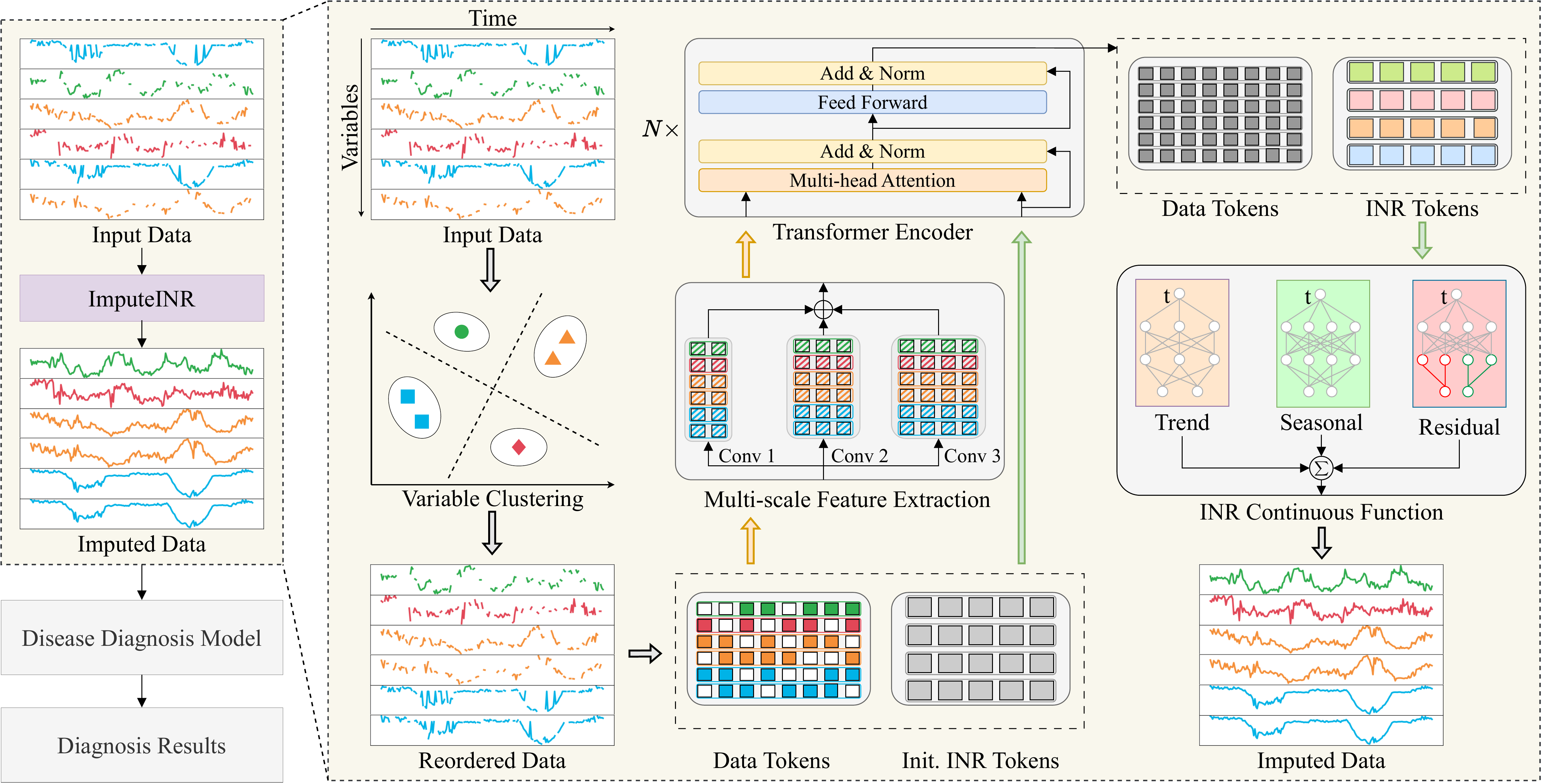}
\caption{The overall workflow of the proposed method. The input data with missing values is imputed using the ImputeINR model, and the imputed data is then fed into the disease diagnosis model to obtain the diagnostic results. In ImputeINR, the INR tokens are predicted using a transformer encoder. These tokens serve as the parameters for the INR continuous function, which takes the timestamp $t$ as input.}
\label{fig_workflow}
\vspace{-1em}
\end{figure*}

\subsection{Implicit Neural Representations on Time Series Data}
Several works have explored the use of INR to model time series data. INRAD \cite{jeong2022time-inrad} leverages INR to reconstruct time series data for anomaly detection. TSINR \cite{li2024tsinr} takes advantage of the spectral bias property of INR, prioritizing low-frequency signals and exhibiting reduced performance on high-frequency anomalous data to identify anomalies. However, these methods are specifically designed for anomaly detection tasks and are not suitable for imputation tasks. Currently, only a few works have attempted to leverage INR for time series imputation tasks, but each of these approaches has its own limitations. For example, HyperTime \cite{fons2022hypertime} utilizes INR to learn a compressed latent representation to capture the underlying patterns in time series for imputation and generation. However, it uses a permutation-invariant set encoder to extract features, leading to an insufficient representation of the underlying patterns in the time series. In addition, TimeFlow \cite{naour2023timeflow} utilizes continuous-time-dependent modeling and INR enhanced by a meta-learning-driven modulation mechanism for imputation and forecasting. However, treats each variable as an individual sample, thereby ignoring crucial inter-variable correlations. Moreover, it fits the entire time series at once becoming computationally inefficient, particularly when dealing with long time series.

\begin{figure*}[t]
\centering
\includegraphics[width = 1\textwidth]{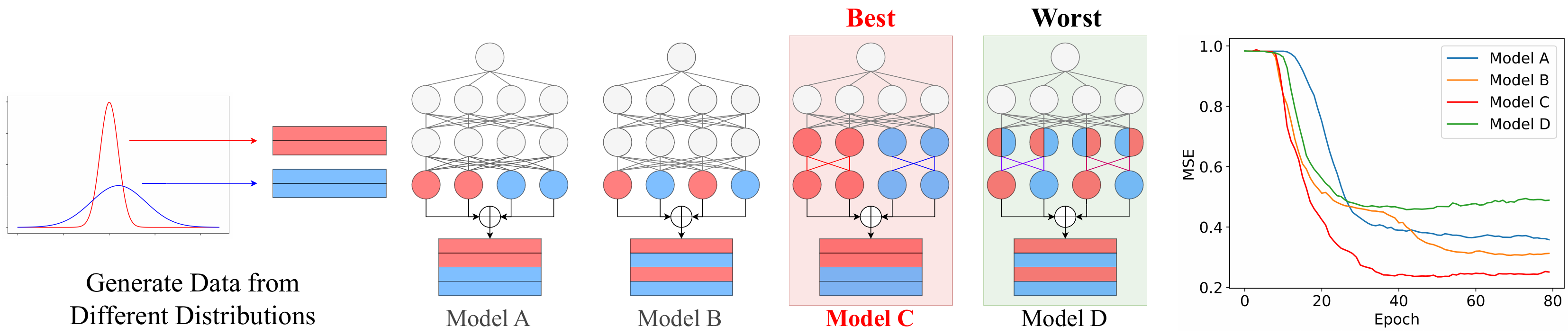}
\caption{The four architectures we test to evaluate the representation capability of the INR continuous function for the synthetic time series dataset. The synthetic dataset consists of four variables, generated from two distinct distributions. The results prove that the representation capability of INR is strongest when variables from the same distribution are fitted by the same group.}
\label{app_fig_motivation}
\vspace{-1em}
\end{figure*}

\section{Methodology}
\subsection{Problem Formulation}
\subsubsection{Imputation with Missing Data}
Denote time series data with $N$ variables and $T$ timestamps as $\mathbf{X}=\left\{\mathbf{x}_1, \mathbf{x}_2, \ldots, \mathbf{x}_N\right\} \in \mathbb{R}^{N \times T}$. The time series data $\mathbf{X}$ is incomplete and the mask rate is $r \in [0,1]$. The corresponding binary mask matrix can be defined as $\mathbf{M}=\left\{m_{n, t}\right\} \in\{0,1\}^{N \times T}$, where $m_{n, t}=1$ if $x_{n, t}$ is observed, and $m_{n, t}=0$ if $x_{n, t}$ is missing. The imputation task is to predict the missing values $\mathbf{X}_{\text{miss}}$ such that the predicted values ${\hat{\mathbf{X}}}$ satisfy ${\hat{\mathbf{X}}}=F_\theta(\mathbf{X}, \mathbf{M})$, where $F_\theta$ mentions the model with parameters $\theta$. The goal is to minimize the reconstruction error between the masked data and the imputed data:
\begin{equation}
    \mathcal{L}({\hat{\mathbf{X}}}, \mathbf{X}_{\text{gt}})=\frac{1}{\left|\mathbf{M}_{\text{miss}}\right|} \sum_{n=1}^N \sum_{t=1}^T\left(1-m_{n, t}\right) \cdot\left(\hat{x}_{n, t}-x_{n, t}\right)^2,
\label{eq_loss}
\end{equation}
where $|\mathbf{M}_{\text{miss}}|$ is the total number of missing values in $\mathbf{X}$ and $\mathbf{X}_{\text{gt}}$ is the ground truth.

\subsubsection{Disease Diagnosis}
Let $\mathbf{X}_{\text{imputed}}$ represent the imputed version of $\mathbf{X}$, where the missing values in $\mathbf{X}$ are imputed and the non-missing values in $\mathbf{X}$ are retained. The task is to utilize $\mathbf{X}_{\text{imputed}}$ for disease diagnosis, where a model $G(\cdot)$ is trained to predict disease labels $\hat{y} \in \{0,1\}$ based on $\mathbf{X}_{\text{imputed}}$:
\begin{equation}
    \hat{y}=G\left(\mathbf{X}_{\text {imputed }}\right),
\end{equation}
where $\hat{y}=0$ denotes normal, and $\hat{y}=1$ indicates the presence of disease.

\subsection{Method Overview}
The core idea of ImputeINR is to leverage the inherent capability of INR to learn continuous functions, allowing for fine-grained imputation of time series data by querying at arbitrary timestamps. By leveraging this flexibility, ImputeINR can generate smooth and accurate estimates for missing data points, even with extremely sparse observed data. However, since time series data has inherently intricate temporal patterns and multi-variable properties, using a simple MLP as the INR continuous function to fit it is challenging. To address these issues, we design a novel form of INR continuous function specifically for time series data. Following the previous work \cite{fons2022hypertime,li2024tsinr}, this form includes three components to capture the trend, seasonal, and residual information to deal with the unique temporal patterns. Furthermore, to enhance the ability of ImputeINR to model multi-variable data, we propose an adaptive group-based architecture to learn complicated residual information. Each group focuses on variables with similar distributions. And we use a clustering algorithm to determine the variable partition. To further enhance the imputation capability of ImputeINR, we incorporate a multi-scale module to capture information at different scales, improving fine-grained imputation performance.

Figure \ref{fig_workflow} demonstrates the overall workflow of the proposed method. The masked data is first reordered based on the variable clustering results so that variables with similar distributions are placed adjacent to each other. This is to enable the subsequent representation of variables within the same cluster using the same group-based MLP in the INR continuous function. Then the reordered masked data is standardized and segmented into patches to prepare the data tokens. Simultaneously, we initialize the INR tokens, which are learnable vector parameters. The processed data tokens are input into convolutional layers of different scales to extract multi-scale features. Subsequently, these extracted features and the initialized INR tokens are fed together into the transformer encoder to predict the INR tokens. These INR tokens are essentially the parameters of the INR continuous function. Based on these parameters, the INR continuous function takes the timestamp $t$ as input and predicts the masked values. Finally, the imputed data is then fed into the disease diagnosis model to obtain the diagnostic results.

\subsection{Variable Clustering}
We adopt a clustering algorithm $\mathcal{C}$ to cluster the variables of the time series data $\mathbf{X}\in \mathbb{R}^{N \times T}$ based on the similarity matrix $S \in \mathbb{R}^{N\times N}$, which partitions the variables into $K$ clusters:
\begin{equation}
    \mathcal{C}: \mathbb{R}^{N\times N} \rightarrow\left\{C_1, C_2, \ldots, C_K\right\},
\end{equation}
where $C_k$ is a subset of the total variable set $\left\{\mathbf{x}_1, \mathbf{x}_2, \ldots, \mathbf{x}_N\right\}$ and its cardinality $\left|C_k\right|$ denotes the number of variables in this cluster. The objective of the clustering function $\mathcal{C}$ is defined as follows:
\begin{equation}
    \operatorname{argmax}_{\left\{C_1, C_2, \ldots, C_K\right\}} \sum_{k=1}^K \sum_{\mathbf{x}_i, \mathbf{x}_j \in C_k} S\left(\mathbf{x}_i, \mathbf{x}_j\right),
\end{equation}
where $S(\mathbf{x}_i,\mathbf{x}_j)$ represents the similarity between variables $\mathbf{x}_i$ and $\mathbf{x}_j$. Then we obtain the permutation matrix $P \in \mathbb{R}^{N \times N}$:
\begin{equation}
    P_{i j}= \begin{cases}1, & \text { if } j=\pi(i), \\ 0, & \text { otherwise, }\end{cases}
\end{equation}
where $\pi$ is the permutation vector that orders the variables according to the clusters. Finally, the reordered matrix $\mathbf{X^\prime}$ with columns permuted according to $\pi$ is given by:
\begin{equation}
    \mathbf{X^{\prime}}=\mathbf{X} \cdot P.
\end{equation}
In this reordered matrix $\mathbf{X^{\prime}}$, rows (i.e., variables) are grouped according to the clusters.

\subsection{Multi-scale Feature Extraction}
To further capture features from different scales for fine-grained imputation, the reordered data $\mathbf{X}^{\prime}\in \mathbb{R}^{N \times T}$ is fed to multiple convolutional layers with varying scales. Each convolutional layer $l$ refers to kernel size $\mathit{k}_l$, stride $\mathit{s}_l$, padding $\mathit{p}_l$, and the number of output channels $\mathit{c}_l$. For each output channel $i$ in the $l^{th}$ convolutional layer, the convolution operation can be formulated as:
\begin{equation}
    \Phi_l\left(\mathbf{X}^{\prime}\right)_{i, t}=\sum_{j=1}^{\mathit{k}_l} W_{l, i, j} \cdot{\mathbf{X}^{\prime}}_{t+j-\mathit{p}_l}+b_{l, i},
\end{equation}
where $W$ and $b$ denote the weight matrix and bias matrix respectively. Then these features of different scales $\Phi_l\left(\mathbf{X}^{\prime}\right) \in \mathbb{R}^{\mathit{c}_l \times\left(T-\mathit{k}_l+2 \mathit{p}_l+1\right)}$ are concatenated to obtained the multi-scale features $\dot{\mathbf{X}} \in \mathbb{R}^{\sum_{l=1}^L \mathit{c}_l \times\left(T^{\prime}-\mathit{k}_l+2 \mathit{p}_l+1\right)}$. Finally, these features are fed to the transformer encoder together with the initialized INR tokens to predict the INR tokens.

\subsection{INR Continuous Function}
INR continuous function $f$ maps the timestamp $t$ to time series data:
\begin{equation}
    f: t \in \mathbb{R} \mapsto \mathbf{X}(t) \in \mathbb{R}^N,
\end{equation}
where $\mathbf{X}(t)$ represents the output values of $N$ variables at timestamp $t$. To effectively capture the complicated temporal patterns and successfully model the multiple variables, we design a novel form of INR continuous function. Following the previous work \cite{fons2022hypertime,li2024tsinr}, our INR continuous function includes three components to model trend, seasonal, and residual patterns separately. It can be defined as follows:
\begin{equation}
    \hat{\mathbf{X}}(t) = f(t) = f_{tre}(t) + f_{sea}(t) + f_{res}(t),
\end{equation}
where $t$ is the input timestamp and $f(t)$ denotes the output (i.e., imputed data). The parameters in the INR continuous function are predicted by the transformer encoder (i.e., INR tokens).

\subsubsection{Trend and Seasonal}
The trend represents the long-term movement or direction of the time series data. It is typically smooth and reflects gradual shifts in the level of the time series, free from noise or short-term fluctuations. Mathematically, it can be modeled as a polynomial function:
\begin{equation}
    f_{tre}(t)=\sum_{i=0}^m \alpha_i t^i,
\end{equation}
where $\alpha_i$ denotes the coefficients and $m$ refers to the degree of the polynomial. In addition, the seasonal component focuses on the repeating patterns or cycles in the time series data, representing predictable fluctuations due to seasonality or recurring events. These regular, cyclical, and short-term fluctuations are modeled with a periodic function:
\begin{equation}
    f_{sea}(t)=\sum_{i=1}^{\lfloor T / 2-1\rfloor}\left(\beta_i \sin \left({2 \pi i t}\right)+\gamma_{i+\lfloor T / 2\rfloor} \cos \left({2 \pi i t}\right)\right),
\end{equation}
where $\beta_i$ and $\gamma_i$ are Fourier coefficients.

\subsubsection{Residual: Adaptive Group-based Architecture}
The residual component represents the unexplained variation after removing the trend and seasonal effects, often modeled as a stochastic process. It is challenging to capture this complex information. As discussed in Figure \ref{app_fig_motivation}, we find that regardless of the order of the variables, using a single MLP is not effective in modeling multiple variables from different distributions. However, if variables from the same distribution are represented using the same set of MLP layers, the performance will significantly improve. We define such a set as a group. In addition, the layers in the MLP that extract information across all variables are called global layers, while the layers within groups are referred to as group layers. The number of groups and their outputs are determined by the results of variable clustering, which allows our architecture to adapt to datasets with various characteristics. It is worth noting that when variables with different distributions are in the same group, the representation capability is significantly reduced. This proves the importance of the correlation information between the variables.

Theoretically, for any given timestamp $t$, we design $L_1$ global layers, $L_2$ group layers, and $K$ groups. $K$ is determined by the results of variable clustering. The global layers are given as follows:
\begin{equation}
    h^{(0)}=t,
\end{equation}
\begin{equation}
    h^{(l_1)}=\sigma\left(W^{(l_1)} h^{(l_1-1)}+b^{(l_1)}\right),
\end{equation}
where $l_1 \in [1, L_1]$, $h^{(l_1)}$ is the output of the $l_{1}^{th}$ global layer, $W$ and $b$ are weight matrix and bias matrix. Then, for group $g_k$, the input is the output of the last global layer:
\begin{equation}
\hat{x}_{g_k}^{(0)}=h^{\left(L_1\right)},
\end{equation}
\begin{equation}
\hat{x}_{g_k}^{(l_2)}=\sigma\left(W_{g_k}^{(l_2)} \hat{x}_{g_k}^{(l_2-1)}+b_{g_k}^{(l_2)}\right),
\end{equation}
where $l_2 \in [1,L_2]$, $\hat{x}_{g_k}^{l_2}$ refers to the output of the $l_2^{th}$ group layer in group $g_k$, $W$ and $b$ are weight matrix and bias matrix. $\hat{x}_{g_k}^{L_2} \in \mathbb{R}^{\left|C_k\right|}$ and $\left|C_k\right|$ is the number of variables in the $k^{th}$ cluster. The final output is the concatenation of the outputs from the last group layer of each group:
\begin{equation}
    f_{res}(t)=\hat{x}_{g_1}^{\left(L_2\right)} \oplus \hat{x}_{g_2}^{\left(L_2\right)} \oplus \ldots \oplus \hat{x}_{g_K}^{\left(L_2\right)}.
\end{equation}

\begin{table*}[!t]
\centering
\caption{Imputation results. The best results are in \textbf{Bold}. And the second ones are \underline{underlined}.}
\vspace{-1em}
\begin{center}
\begin{small}
\scalebox{0.865}{
\setlength\tabcolsep{3pt}
\begin{tabular}{cc|cc|cc|cc|cc|cc|cc|cc|cc|cc|cc}
\toprule
\multicolumn{2}{c|}{Methods}                  & \multicolumn{2}{c|}{ImputeINR}   & \multicolumn{2}{c|}{ImputeFormer} & \multicolumn{2}{c|}{TimeMixer} & \multicolumn{2}{c|}{iTransformer} & \multicolumn{2}{c|}{FPT}         & \multicolumn{2}{c|}{TimesNet} & \multicolumn{2}{c|}{SAITS} & \multicolumn{2}{c|}{BRITS}    & \multicolumn{2}{c|}{SSSD}  & \multicolumn{2}{c}{CSDI}     \\
\multicolumn{2}{c|}{Mask Rate}                & MSE            & MAE            & MSE              & MAE           & MSE              & MAE        & MSE             & MAE            & MSE            & MAE            & MSE           & MAE          & MSE         & MAE         & MSE         & MAE            & MSE         & MAE         & MSE         & MAE            \\ \midrule
\multirow{5}{*}{\rotatebox{90}{ETT}} & 10\% & 0.020 & 0.098          & 0.021            & 0.091         & 0.035            & 0.115      & 0.042           & 0.141          & \textbf{0.017} & \textbf{0.087} & {\ul 0.018}   & {\ul 0.088}  & 0.021       & 0.100       & 0.021       & 0.089          & 0.022       & 0.098       & 0.019       & \textbf{0.087} \\
                              & 30\% & 0.027 & 0.109          & \textbf{0.023}   & {\ul 0.098}   & 0.041            & 0.125      & 0.066           & 0.180          & 0.030          & 0.110          & 0.031         & 0.111        & 0.030       & 0.114       & 0.028       & 0.110          & 0.027       & 0.109       & {\ul 0.024} & \textbf{0.097} \\
                              & 50\%          & \textbf{0.028} & \textbf{0.111} & 0.034            & 0.116         & 0.054            & 0.143      & 0.109           & 0.234          & 0.041          & 0.130          & 0.035         & 0.123        & 0.031       & 0.116       & 0.040       & 0.130          & {\ul 0.029} & {\ul 0.112} & 0.034       & {\ul 0.112}    \\
                              & 70\%          & \textbf{0.039} & \textbf{0.134} & 0.050            & 0.142         & 0.077            & 0.170      & 0.124           & 0.246          & 0.085          & 0.181          & 0.057         & 0.155        & {\ul 0.043} & {\ul 0.135} & 0.068       & 0.181          & 0.044       & {\ul 0.135} & 0.049       & {\ul 0.135}    \\
                              & 90\%          & \textbf{0.095} & \textbf{0.214} & {\ul 0.122}      & 0.218         & 0.223            & 0.276      & 0.247           & 0.336          & 0.272          & 0.309          & 0.231         & 0.295        & 0.213       & 0.218       & 0.251       & 0.358          & 0.154       & 0.254       & 0.124       & {\ul 0.216}    \\ \midrule
\multirow{5}{*}{\rotatebox{90}{Weather}}      & 10\%          & \textbf{0.026} & \textbf{0.063} & 0.032            & 0.076         & 0.029            & 0.069      & 0.036           & 0.081          & 0.028          & {\ul 0.064}    & 0.028         & {\ul 0.064}  & 0.031       & 0.073       & {\ul 0.027} & \textbf{0.063} & 0.036       & 0.069       & 0.039       & 0.065          \\
                              & 30\%          & \textbf{0.030} & \textbf{0.072} & 0.033            & 0.080         & 0.032            & 0.080      & 0.051           & 0.113          & 0.035          & 0.075          & {\ul 0.031}   & {\ul 0.073}  & 0.035       & 0.077       & {\ul 0.031} & {\ul 0.073}    & 0.039       & {\ul 0.073} & 0.043       & 0.074          \\
                              & 50\%          & \textbf{0.031} & \textbf{0.073} & 0.037            & 0.084         & 0.037            & 0.076      & 0.069           & 0.144          & 0.043          & 0.076          & 0.036         & 0.076        & 0.041       & 0.091       & {\ul 0.035} & 0.077          & 0.040       & {\ul 0.075} & 0.048       & 0.076          \\
                              & 70\%          & \textbf{0.036} & \textbf{0.082} & 0.074            & 0.097         & 0.045            & 0.086      & 0.078           & 0.147          & 0.053          & 0.087          & 0.043         & {\ul 0.084}  & 0.047       & 0.096       & {\ul 0.042} & 0.085          & 0.053       & 0.086       & 0.057       & 0.093          \\
                              & 90\%          & \textbf{0.065} & \textbf{0.123} & 0.082            & 0.126         & 0.076            & 0.126      & {\ul 0.124}     & 0.191          & 0.089          & 0.129          & 0.073         & 0.125        & {\ul 0.066} & {\ul 0.124} & 0.090       & 0.130          & 0.089       & 0.128       & 0.088       & 0.127          \\ \midrule
\multirow{5}{*}{\rotatebox{90}{BAQ}}          & 10\%          & \textbf{0.083} & \textbf{0.169} & 1.050            & 0.747         & {\ul 0.165}      & 0.172      & 0.235           & 0.258          & 0.215          & 0.224          & 0.262         & 0.266        & 1.085       & 0.748       & 0.208       & 0.175          & 0.306       & 0.357       & 0.201       & {\ul 0.171}    \\
                              & 30\%          & \textbf{0.096} & \textbf{0.171} & 1.096            & 0.749         & 0.205            & 0.193      & 0.308           & 0.321          & 0.231          & 0.229          & 0.292         & 0.267        & 1.088       & 0.749       & 0.210       & 0.186          & 0.368       & 0.365       & {\ul 0.203} & {\ul 0.174}    \\
                              & 50\%          & \textbf{0.101} & \textbf{0.172} & 1.106            & 0.750         & 0.274            & 0.237      & 0.404           & 0.399          & 0.285          & 0.242          & 0.318         & 0.269        & 1.112       & 0.750       & 0.211       & 0.191          & 0.392       & 0.401       & {\ul 0.209} & {\ul 0.183}    \\
                              & 70\%          & \textbf{0.117} & \textbf{0.181} & 1.119            & 0.751         & 0.359            & 0.289      & 0.556           & 0.488          & 0.325          & 0.262          & 0.341         & 0.280        & 1.124       & 0.751       & 0.230       & 0.206          & 0.462       & 0.424       & {\ul 0.229} & {\ul 0.192}    \\
                              & 90\%          & \textbf{0.122} & \textbf{0.185} & 1.129            & 0.752         & 0.503            & 0.367      & 0.803           & 0.615          & 0.430          & 0.301          & 0.427         & 0.317        & 1.127       & 0.752       & 0.411       & 0.308          & 0.671       & 0.532       & {\ul 0.322} & {\ul 0.218}    \\ \midrule
\multirow{5}{*}{\rotatebox{90}{IAQ}}          & 10\%          & \textbf{0.007} & \textbf{0.061} & 1.340            & 0.725         & 0.139            & 0.171      & 0.592           & 0.466          & 0.228          & 0.264          & 0.248         & 0.286        & 1.277       & 0.735       & 0.164       & 0.210          & 0.144       & 0.185       & {\ul 0.064} & {\ul 0.116}    \\
                              & 30\%          & \textbf{0.008} & \textbf{0.062} & 1.377            & 0.738         & 0.244            & 0.242      & 0.639           & 0.503          & 0.237          & 0.271          & 0.262         & 0.290        & 1.442       & 0.755       & 0.224       & 0.243          & 0.199       & 0.192       & {\ul 0.074} & {\ul 0.122}    \\
                              & 50\%          & \textbf{0.009} & \textbf{0.063} & 1.424            & 0.753         & 0.375            & 0.306      & 0.783           & 0.556          & 0.291          & 0.305          & 0.274         & 0.297        & 1.461       & 0.757       & 0.241       & 0.273          & 0.213       & 0.216       & {\ul 0.101} & {\ul 0.144}    \\
                              & 70\%          & \textbf{0.010} & \textbf{0.068} & 1.466            & 0.757         & 0.527            & 0.377      & 0.907           & 0.618          & 0.426          & 0.357          & 0.304         & 0.314        & 1.472       & 0.761       & 0.504       & 0.355          & 0.343       & 0.280       & {\ul 0.238} & {\ul 0.225}    \\
                              & 90\%          & \textbf{0.029} & \textbf{0.116} & 1.478            & 0.761         & 0.847            & 0.498      & 1.205           & 0.767          & 0.811          & 0.504          & 0.720         & 0.477        & 1.493       & 0.764       & 0.981       & 0.505          & 0.941       & 0.523       & {\ul 0.657} & {\ul 0.428}    \\ \midrule
\multirow{5}{*}{\rotatebox{90}{Solar}}        & 10\%          & \textbf{0.022} & \textbf{0.074} & 0.768            & 0.771         & 0.024            & 0.079      & 0.060           & 0.167          & {\ul 0.075}    & 0.173          & 0.048         & 0.132        & 0.770       & 0.772       & {\ul 0.023} & {\ul 0.075}    & 0.122       & 0.143       & {\ul 0.023} & {\ul 0.075}    \\
                              & 30\%          & \textbf{0.023} & \textbf{0.075} & 0.770            & 0.772         & 0.034            & 0.107      & 0.071           & 0.181          & 0.084          & 0.185          & 0.049         & 0.133        & 0.771       & 0.773       & {\ul 0.024} & {\ul 0.076}    & 0.128       & 0.147       & 0.025       & 0.078          \\
                              & 50\%          & \textbf{0.024} & \textbf{0.078} & 0.772            & 0.773         & 0.052            & 0.143      & 0.079           & 0.189          & 0.101          & 0.202          & 0.052         & 0.139        & 0.772       & 0.774       & {\ul 0.026} & {\ul 0.080}    & 0.145       & 0.159       & {\ul 0.026} & {\ul 0.080}    \\
                              & 70\%          & \textbf{0.025} & \textbf{0.079} & 0.773            & 0.774         & 0.075            & 0.173      & 0.088           & 0.200          & 0.139          & 0.243          & 0.061         & 0.151        & 0.773       & 0.775       & {\ul 0.030} & 0.085          & 0.211       & 0.228       & 0.036       & {\ul 0.084}    \\
                              & 90\%          & \textbf{0.026} & \textbf{0.081} & 0.774            & 0.775         & 0.166            & 0.249      & 0.120           & 0.250          & 0.435          & 0.444          & 0.121         & 0.211        & 0.774       & 0.776       & 0.052       & 0.100          & 0.426       & 0.312       & {\ul 0.041} & {\ul 0.085}    \\ \midrule
\multirow{5}{*}{\rotatebox{90}{Phy2012}}      & 10\%          & \textbf{0.072} & \textbf{0.096} & 0.200            & 0.153         & 0.104            & 0.115      & 0.097           & 0.108          & 0.087          & 0.104          & {\ul 0.080}   & 0.101        & 0.200       & 0.163       & 0.097       & 0.100          & {\ul 0.080} & {\ul 0.098} & 0.656       & 0.514          \\
                              & 30\%          & \textbf{0.079} & \textbf{0.101} & 0.205            & 0.155         & 0.117            & 0.120      & {\ul 0.099}     & 0.111          & {\ul 0.099}    & 0.111          & 0.103         & 0.108        & 0.203       & 0.168       & 0.108       & 0.105          & 0.107       & {\ul 0.102} & 0.828       & 0.585          \\
                              & 50\%          & \textbf{0.092} & \textbf{0.107} & 0.210            & 0.158         & 0.142            & 0.124      & 0.109           & 0.115          & {\ul 0.105}    & 0.118          & 0.145         & 0.118        & 0.208       & 0.173       & 0.117       & 0.116          & 0.139       & {\ul 0.108} & 0.923       & 0.601          \\
                              & 70\%          & \textbf{0.071} & \textbf{0.112} & 0.229            & 0.169         & 0.148            & 0.129      & {\ul 0.124}     & 0.120          & 0.132          & 0.131          & 0.149         & 0.128        & 0.237       & 0.195       & 0.125       & 0.123          & 0.149       & {\ul 0.118} & 0.994       & 0.670          \\
                              & 90\%          & \textbf{0.127} & \textbf{0.124} & 0.232            & 0.170         & 0.179            & 0.143      & {\ul 0.160}     & {\ul 0.135}    & 0.167          & 0.145          & 0.177         & 0.144        & 0.214       & 0.159       & 0.163       & 0.139          & 0.226       & 0.157       & 1.068       & 0.705          \\ \midrule
\multirow{5}{*}{\rotatebox{90}{Phy2019}}      & 10\%          & \textbf{0.071} & \textbf{0.102} & 0.199            & 0.159         & 0.100            & 0.116      & {\ul 0.072}     & 0.104          & 0.082          & 0.111          & 0.075         & 0.105        & 0.199       & 0.168       & 0.089       & {\ul 0.103}    & 0.083       & 0.109       & 0.718       & 0.579          \\
                              & 30\%          & \textbf{0.079} & \textbf{0.109} & 0.206            & 0.160         & 0.104            & 0.120      & 0.098           & 0.122          & 0.091          & 0.116          & {\ul 0.084}   & 0.111        & 0.203       & 0.169       & 0.099       & {\ul 0.110}    & 0.085       & 0.114       & 0.813       & 0.648          \\
                              & 50\%          & \textbf{0.087} & \textbf{0.115} & 0.209            & 0.164         & 0.109            & 0.125      & 0.100           & 0.123          & 0.102          & 0.124          & {\ul 0.094}   & {\ul 0.118}  & 0.204       & 0.175       & 0.109       & {\ul 0.118}    & 0.098       & {\ul 0.118} & 0.956       & 0.657          \\
                              & 70\%          & \textbf{0.098} & \textbf{0.120} & 0.211            & 0.172         & 0.119            & 0.132      & 0.112           & 0.129          & 0.116          & 0.133          & {\ul 0.109}   & 0.128        & 0.205       & 0.178       & 0.122       & {\ul 0.124}    & 0.167       & 0.142       & 0.985       & 0.673          \\
                              & 90\%          & \textbf{0.121} & \textbf{0.131} & 0.214            & 0.174         & 0.152            & 0.149      & {\ul 0.123}     & {\ul 0.132}    & 0.153          & 0.152          & 0.149         & 0.149        & 0.206       & 0.180       & 0.151       & 0.142          & 0.226       & 0.173       & 1.061       & 0.761          \\ \midrule
\multirow{5}{*}{\rotatebox{90}{MIMIC3}}       & 10\%          & \textbf{0.019} & \textbf{0.041} & 0.149            & 0.142         & 0.049            & 0.077      & 0.052           & 0.085          & {\ul 0.031}    & 0.055          & 0.043         & 0.070        & 0.150       & 0.148       & {\ul 0.031} & {\ul 0.045}    & 0.036       & {\ul 0.045} & 0.096       & 0.067          \\
                              & 30\%          & \textbf{0.023} & \textbf{0.044} & 0.150            & 0.143         & 0.081            & 0.079      & 0.055           & 0.089          & {\ul 0.033}    & 0.057          & 0.051         & 0.072        & 0.151       & 0.149       & 0.039       & {\ul 0.046}    & 0.094       & 0.050       & 0.157       & 0.084          \\
                              & 50\%          & \textbf{0.027} & \textbf{0.048} & 0.151            & 0.144         & 0.075            & 0.082      & 0.061           & 0.100          & {\ul 0.040}    & 0.063          & 0.076         & 0.074        & 0.152       & 0.150       & 0.066       & {\ul 0.057}    & 0.102       & 0.058       & 0.216       & 0.088          \\
                              & 70\%          & \textbf{0.033} & \textbf{0.056} & 0.161            & 0.145         & 0.083            & 0.086      & 0.084           & 0.113          & {\ul 0.081}    & 0.068          & 0.093         & 0.080        & 0.155       & 0.153       & 0.083       & {\ul 0.059}    & 0.110       & 0.064       & 0.220       & 0.114          \\
                              & 90\%          & \textbf{0.062} & \textbf{0.070} & 0.178            & 0.148         & 0.126            & 0.097      & 0.136           & 0.130          & {\ul 0.098}    & 0.080          & 0.134         & 0.094        & 0.169       & 0.156       & 0.112       & {\ul 0.077}    & 0.128       & 0.078       & 0.354       & 0.201          \\ \midrule
\multicolumn{2}{c|}{Average}             & \textbf{0.054} & \textbf{0.102} & 0.496            & 0.371         & 0.158            & 0.164      & 0.232           & 0.238          & 0.161          & 0.176          & 0.148         & 0.166        & 0.499       & 0.376       & {\ul 0.142} & {\ul 0.146}    & 0.186       & 0.178       & 0.325       & 0.260          \\ \bottomrule
\end{tabular}}
\vspace{-1em}
\end{small}
\end{center}
\label{table_main_results}
\end{table*}

\section{Experiments}
\subsection{Experimental Setup}
\subsubsection{Datasets and Baseline Methods}
We use eight time series imputation benchmark datasets to validate the performance of ImputeINR, including ETT \cite{zhou2021informer-ett}, Weather \cite{weather}, BAQ \cite{zhang2017airquality-baq}, IAQ \cite{vito2016air-iaq}, Solar \cite{solar}, Phy2012 \cite{silva2012physionet-phy2012}, Phy2019 \cite{reyna2019physionet-phy2019}), and MIMIC3 \cite{johnson2018mimic}. Further, we select the Phy2012, Phy2019, and MIMIC3 healthcare datasets to validate the effectiveness of the imputation results produced by ImputeINR for downstream disease diagnosis tasks. To verify the superiority of ImputeINR, we compare our method to nine state-of-the-art imputation methods, including RNN-based methods (BRITS \cite{cao2018brits}), CNN-based methods (TimesNet \cite{wutimesnet}), MLP-based methods (TimeMixer \cite{wangtimemixer}), transformer-based methods (SAITS \cite{du2023saits}, FPT \cite{zhou2023one-fpt}, iTransformer \cite{liuitransformer}, ImputeFormer \cite{nie2024imputeformer}), and diffusion-based methods (CSDI \cite{tashiro2021csdi}, SSSD \cite{alcaraz2022diffusion-sssd}).

\subsubsection{Experimental Settings}
We apply the same data processing techniques and parameter settings. A sliding window approach is used, with a fixed window size of 48 for the Phy2012, Phy2019, and MIMIC3 datasets, and 96 for all other datasets. These settings follow those used in previous work \cite{wutimesnet,du2023pypots}. To evaluate the imputation performance, we use the same masking strategy as previous works \cite{wutimesnet,zhou2023one-fpt}, which randomly mask values in $\mathbf{X}_{\mathrm{gt}}$ based on the mask rate $r$. For the imputation results, the multi-scale feature extraction module uses three parallel convolutional layers with kernel sizes of 3,5,7 respectively. The adaptive group-based architecture in the INR continuous function involves one global layer and one group layer within the residual component, with hidden dimensions set to 16. The transformer encoder consists of 6 blocks. We use the agglomerative clustering method to achieve variable clustering since it adopts diverse inputs without the need to pre-specify the number of clusters. Ablation Studies are reported in Section \ref{ablation_studies} to demonstrate the effectiveness of each module. For the downstream disease diagnosis task, we employ the default Long Short-Term Memory (LSTM) classifier provided by the official benchmark \cite{Harutyunyan2019-lstm} as the disease diagnosis model, with the diagnostic results presented in Section \ref{disease_diagnosis}. Experiments are performed using the ADAM optimizer \cite{kingma2014adam} with an initial learning rate of $10^{-3}$. All experiments are conducted on a single 24GB GeForce RTX 3090 GPU. The efficiency analysis is in Section \ref{efficiency_analysis}.

\begin{table*}[!t]
\centering
\caption{The ablation studies of each module in ImputeINR. The best results are in \textbf{Bold}.}
\vspace{-1em}
\begin{center}
\begin{small}
\scalebox{0.9}{
\setlength\tabcolsep{3pt}
\begin{tabulary}{0.5cm}{ccc|cc|cc|cc|cc|cc|cc|cc|cc}
\toprule
Multi-scale & Variable   & Adaptive & \multicolumn{2}{c|}{ETT}         & \multicolumn{2}{c|}{Weather}     & \multicolumn{2}{c|}{BAQ}         & \multicolumn{2}{c|}{IAQ}         & \multicolumn{2}{c|}{Solar}       & \multicolumn{2}{c|}{Phy2012}     & \multicolumn{2}{c|}{Phy2019}     & \multicolumn{2}{c}{MIMIC3}      \\
Features    & Clustering & Group    & MSE            & MAE            & MSE            & MAE            & MSE            & MAE            & MSE            & MAE            & MSE            & MAE            & MSE            & MAE            & MSE            & MAE            & MSE            & MAE            \\ \midrule
\ding{55}           & \ding{55}          & \ding{55}        & 0.039          & 0.135          & 0.038          & 0.083          & 0.227          & 0.262          & 0.018          & 0.092          & 0.036          & 0.106          & 0.099          & 0.114          & 0.098          & 0.119          & 0.038          & 0.062          \\ \midrule
\ding{55}           & \ding{55}          & \ding{51}        & 0.036          & 0.130          & 0.035          & 0.081          & 0.222          & 0.258          & 0.015          & 0.084          & 0.034          & 0.098          & 0.099          & 0.113          & 0.096          & 0.117          & 0.036          & 0.056          \\
\ding{55}           & \ding{51}          & \ding{55}        & 0.036          & 0.129          & 0.036          & 0.080          & 0.218          & 0.259          & 0.015          & 0.083          & 0.033          & 0.096          & 0.098          & 0.113          & 0.097          & 0.118          & 0.035          & 0.057          \\
\ding{51}           & \ding{55}          & \ding{55}        & 0.035          & 0.127          & 0.035          & 0.082          & 0.209          & 0.252          & 0.017          & 0.088          & 0.033          & 0.100          & 0.097          & 0.114          & 0.095          & 0.117          & 0.037          & 0.055          \\ \midrule
\ding{55}           & \ding{51}          & \ding{51}        & 0.029          & 0.115          & 0.032          & 0.074          & 0.192          & 0.243          & 0.010          & 0.066          & 0.031          & 0.092          & 0.093          & 0.108          & 0.088          & 0.111          & 0.028          & 0.049          \\
\ding{51}           & \ding{55}          & \ding{51}        & 0.034          & 0.124          & 0.034          & 0.079          & 0.203          & 0.248          & 0.012          & 0.077          & 0.031          & 0.094          & 0.095          & 0.113          & 0.093          & 0.116          & 0.033          & 0.053          \\
\ding{51}           & \ding{51}          & \ding{55}        & 0.033          & 0.123          & 0.033          & 0.078          & 0.199          & 0.244          & 0.014          & 0.081          & 0.032          & 0.096          & 0.096          & 0.113          & 0.094          & 0.117          & 0.035          & 0.051          \\ \midrule
\ding{51}           & \ding{51}          & \ding{51}        & \textbf{0.028} & \textbf{0.111} & \textbf{0.031} & \textbf{0.073} & \textbf{0.101} & \textbf{0.172} & \textbf{0.009} & \textbf{0.063} & \textbf{0.024} & \textbf{0.078} & \textbf{0.092} & \textbf{0.107} & \textbf{0.087} & \textbf{0.115} & \textbf{0.027} & \textbf{0.048}\\
\bottomrule
\end{tabulary}}
\vspace{-1em}
\end{small}
\end{center}
\label{table_ablation_studies}
\end{table*}

\begin{table*}[!t]
\centering
\caption{The disease diagnosis results. The AUROC values are reported and the best results are in \textbf{Bold}.}
\vspace{-1em}
\begin{center}
\begin{small}
\scalebox{1}{
\setlength\tabcolsep{3pt}
\begin{tabulary}{0.5cm}{c|c|ccccccccccc}
\toprule
Dataset & ImputeINR       & ImputeFormer & TimeMixer & iTransformer & FPT    & TimesNet & SAITS  & BRITS  & SSSD   & CSDI   & Mean   & Zero   \\ \midrule
Phy2012 & \textbf{0.8382} & 0.8238       & 0.8304    & 0.8305       & 0.8230 & 0.8286   & 0.8299 & 0.8289 & 0.8215 & 0.8325 & 0.8175 & 0.8023 \\
Phy2019 & \textbf{0.7346} & 0.7050       & 0.6923    & 0.6737       & 0.7117 & 0.6964   & 0.6743 & 0.6968 & 0.6818 & 0.6980 & 0.7095 & 0.6871 \\
MIMIC3  & \textbf{0.8604} & 0.8482       & 0.8570    & 0.8564       & 0.8540 & 0.8533   & 0.8500 & 0.8527 & 0.8096 & 0.8526 & 0.8472 & 0.8419 \\
\bottomrule
\end{tabulary}}
\vspace{-1.5em}
\end{small}
\end{center}
\label{table_disease_diagnosis}
\end{table*}

\subsection{Imputation Task}
\subsubsection{Main Results}
We compare our ImputeINR method to nine state-of-the-art imputation methods with five different mask rates $r$. As shown in Table \ref{table_main_results}, our ImputeINR achieves the best performance in most conditions in terms of both MSE and MAE metrics. Overall, across all datasets and mask rates, our method achieves an average MSE reduction of 62.0\% compared to the second-best results. These results demonstrate that our proposed ImputeINR can effectively deal with datasets of various sizes. In addition, we observe that the performance of most methods declines as the mask rate $r$ increases. This aligns with our expectations, as fewer samples are captured leading to incomplete information, which increases the difficulty of imputation. However, ImputeINR is still effective even with an extreme mask rate. When 90\% of the data is masked, the average MSE of our method is reduced by 68.2\% compared to the second-best ones. This indicates that ImputeINR can learn continuous functions from very few data points, achieving fine-grained imputation.

\subsubsection{Ablation Studies}
\label{ablation_studies}
In this section, we conduct ablation studies to evaluate the effectiveness of the multi-scale feature extraction block, variable clustering, and adaptive group-based architecture. We set mask rate $r$ to be 50\% for the ablation study, as it provides a moderate level of missingness that effectively highlights the impact of each model component while maintaining stability and representativeness in the results. Table \ref{table_ablation_studies} presents the imputation results for all conditions. First, the model without any of the three modules exhibits the lowest performance. Building on this, adding any one of the modules will enhance the imputation capability of the model. This individually validates the effectiveness of each of the three modules. Furthermore, the permutation of any two modules will lead to higher performance. Among them, the combination of variable clustering and adaptive group-based architecture yields the best results. This is as expected, since the variable clustering determines the variable partition and its outcomes correspond directly to the number of groups. Therefore, these two modules can support each other, facilitating better representational learning. Finally, the model using all three modules displays the highest imputation performance as each module contributes complementary strengths.

\subsection{Downstream Disease Diagnosis Task}
\label{disease_diagnosis}
To evaluate the effectiveness of our proposed ImputeINR method in real-world healthcare applications, we conduct disease diagnosis using the imputed data. Specifically, we first apply ImputeINR to impute missing values in healthcare datasets and then use the imputed data for disease diagnosis. As shown in Table \ref{table_disease_diagnosis}, the performance is compared against commonly used imputation methods, including zero imputation, mean imputation, and other state-of-the-art imputation techniques. For evaluation, we employ the standard metric of area under the receiver operating characteristic curve (AUROC). The results demonstrate that the imputed data from ImputeINR consistently leads to superior diagnostic performance across all metrics. This indicates that our model not only effectively reconstructs missing values but also preserves critical disease-related patterns, enhancing the downstream diagnostic capability.

\begin{figure}[]
\centering
\includegraphics[width = 0.9\linewidth]{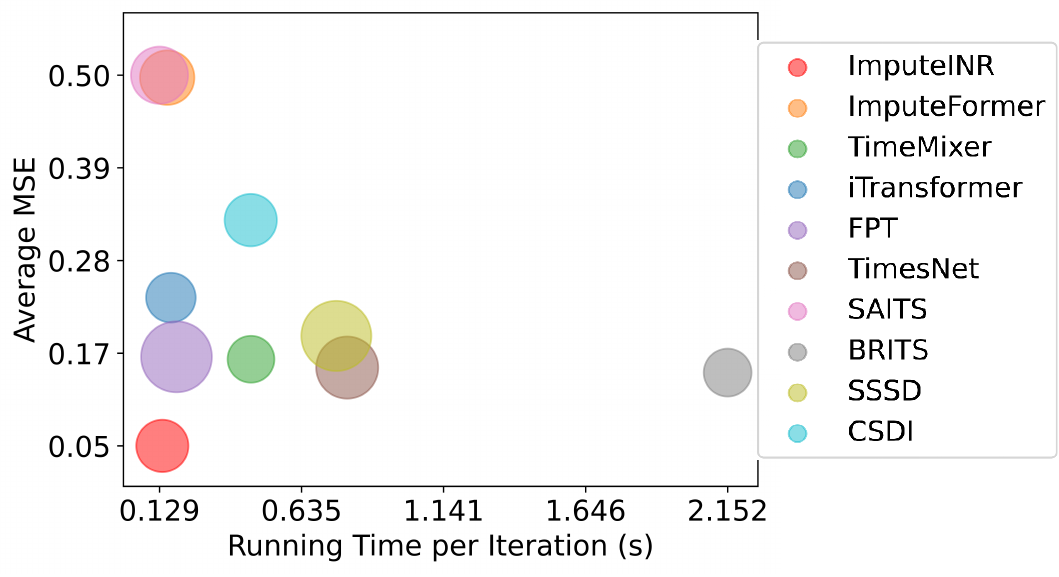}
\vspace{-1em}
\caption{A bubble chart reporting running time vs. imputation performance. The size of each bubble refers to the model size.}
\label{fig_efficiency}
\vspace{-1em}
\end{figure}

\subsection{Efficiency Analysis}
\label{efficiency_analysis}
We evaluate the efficiency of the ImputeINR method, with the results presented in Figure \ref{fig_efficiency}. This bubble chart provides a comprehensive overview of the model performance. On the x-axis, we plot the running time per iteration, while the y-axis represents the average MSE. The size of each bubble is proportional to the number of model parameters, allowing for a direct comparison of model complexity. The bottom-left corner, which corresponds to models with lower running time and better imputation accuracy, indicates superior performance. This figure shows that ImputeINR is closest to the bottom-left corner, which signifies that it achieves accurate imputation results while requiring relatively little running time. Moreover, the corresponding bubble size suggests that ImputeINR has a relatively small number of model parameters, further emphasizing its efficiency. These results prove that ImputeINR strikes a favorable balance between computational efficiency and imputation accuracy.

\section{Conclusion}
In this paper, we propose ImputeINR, which can achieve effective imputation even in extremely sparse scenarios and the imputed data can further enhance the performance of downstream disease diagnosis tasks.  It learns the INR continuous function to map timestamps to the corresponding variable values. In contrast to existing imputation approaches, ImputeINR leverages the sampling frequency-independent and infinite-sampling frequency capabilities of INR to achieve fine-grained imputation with sparse observed data. 

\newpage
\section*{Acknowledgements}
This work was supported by the National Natural Science Foundation of China (Grant Nos. 62202422 and 62372408).

\bibliographystyle{named}
\bibliography{ijcai25}

\newpage
\appendix
\section{Appendix}
\subsection{Data Continuity and Implicit Neural Representations}
The real-world signals are not discrete, but they are represented in a discrete manner. For instance, we represent time series as sequences of discrete points, using sampled values at specific time intervals. However, these discrete representations come with a significant drawback: they only capture an absent amount of information about the signal. Therefore, to utilize this discrete sampling information to represent the complete continuous signal, we need to learn a continuous function $f$ that parameterizes the signal mathematically. With a timestamp $t$, $f$ outputs the corresponding values at that time. And we can sample the time series at any time point from $f$.

This type of continuous function is called INR. INRs are neural networks (e.g., MLPs) that estimate the function $f$ that represents a signal continuously by training on discretely represented samples of the same signal, based on the idea that neural networks can estimate complex functions after observing training data. The process to learn continuous function $f$ is:
\begin{equation}
    f\left(x, \phi, \nabla_x \phi, \nabla_x^2 \phi, \cdots\right)=0, \phi: x \mapsto \phi(x),
\end{equation}
where $\phi$ is parameterized by the network and the estimated $f$ is implicitly encoded in the network after training on the discretely represented samples.

Compared to finite resolution methods, the advantages of INRs become particularly clear. Grid-based representations, such as those used in standard image processing or time series analysis, are inherently tied to a fixed sampling rate. For example, a time series recorded at 10-minute intervals cannot natively provide values at 5-minute or 1-minute intervals without applying an external interpolation method, such as linear or cubic interpolation. Similarly, CNNs used for image reconstruction operate on fixed-resolution tensors; if higher-resolution outputs are desired, these models must be retrained, or separate modules for super-resolution must be added, which can introduce artifacts or limit generalization. Transformer models, which are increasingly used in sequence modeling, also often rely on positional encodings that are resolution-specific, making it difficult for them to handle unseen positions or different temporal granularities.

Unlike traditional representations that use discrete data points (like pixels in images or time points in time series), INRs encode information through functions that map coordinates (such as spatial positions or timestamps) directly to values (like colors or variables). Therefore, INRs allow for smooth representation of data, making it possible to generate high-resolution outputs from low-resolution inputs by querying the model at arbitrary points. And the continuous nature of INRs can lead to better generalization in tasks requiring interpolation or extrapolation of data.

\subsection{Algorithm}
\label{app_algorithm}
The algorithm of variable clustering (Algorithm \ref{app_algo_variable}) and the overall ImputeINR imputation method (Algorithm \ref{app_algo_imputeinr}) are presented as follows. All experiments are implemented based on PyTorch.

For the variable clustering (Algorithm \ref{app_algo_variable}), we employ the agglomerative clustering method because it allows for varied inputs and does not require a predetermined number of clusters. Agglomerative clustering is a hierarchical clustering technique that starts with each data point as its own individual cluster. The algorithm iteratively merges the closest pairs of clusters based on a chosen distance metric until a stopping criterion is met. This method is particularly useful for its flexibility, as it does not require the number of clusters to be specified in advance, making it suitable for exploratory data analysis. With this clustering method, we obtain the clusters $C$ which is used to determine the specific settings of the adaptive group-based architecture.

For the overall ImputeINR imputation method (Algorithm \ref{app_algo_imputeinr}), we predict the masked values with our designed INR continuous function. The masked data is firstly reordered based on the variable clustering results to make variables with similar distributions adjacent and then fed into a multi-scale feature extraction module to capture information from different time scales. The extracted features are entered in a transformer encoder with initialized INR tokens to predict the INR tokens. These learned INR tokens are the parameters of the INR continuous functions. More specifically, these parameters are not learnable but are predicted by the transformer encoder. With the predicted parameters, we input timestamp $t$ to calculate the corresponding variable values as the imputed data. The objective function (i.e., loss function) is the reconstruction error between the masked data and the imputed data as mentioned in Equation \ref{eq_loss}.

\begin{algorithm}
\caption{Variable Clustering}
\label{alg:agglomerative_clustering}
\begin{algorithmic}[1]
\STATE \textbf{Input:} Data points $\mathbf{X} = \{\mathbf{x}_1, \mathbf{x}_2, \ldots, \mathbf{x}_n\}$, distance metric $d$, stopping criterion $\epsilon$
\STATE \textbf{Output:} Clusters $C$
\STATE Initialize each point as its own cluster: $C = \{ \{x_1\}, \{x_2\}, \ldots, \{x_n\} \}$
\WHILE{the number of clusters $|C| > 1$}
    \STATE Find the closest pair of clusters $C_i, C_j$ such that $d(C_i, C_j) = \min_{C_k, C_l \in C} d(C_k, C_l)$
    \IF{$d(C_i, C_j) < \epsilon$}
        \STATE Merge clusters: $C \leftarrow C \setminus \{C_i, C_j\} \cup \{C_i \cup C_j\}$
    \ENDIF
\ENDWHILE
\STATE \textbf{return} $C$
\end{algorithmic}
\label{app_algo_variable}
\end{algorithm}

\begin{algorithm}
\caption{ImputeINR Imputation Algorithm}
\label{alg:inr_interpolation}
\begin{algorithmic}[1]
\STATE \textbf{Input:} Time series data $\mathbf{X}$ with missing values, mask rate $r$
\STATE \textbf{Output:} Imputed values $\hat{\mathbf{X}}$
\STATE Perform feature clustering on the $N$ features of $\mathbf{X}$ to obtain clusters $C$
\STATE Reorder $\mathbf{X}$ based on clusters $C$ to get $\mathbf{X^{\prime}}$
\FOR{each convolutional layer in the multi-scale feature extraction module}
    \STATE Extract features from $\mathbf{X^{\prime}}$ using different kernel sizes
\ENDFOR
\STATE Concatenate outputs from all convolutional layers to obtain $\dot{\mathbf{X}}$
\STATE Initialize INR tokens $\theta$
\STATE Input $\dot{\mathbf{X}}$ and $\theta$ into transformer encoder to predict the INR tokens $\theta^{*}$
\FOR{each timestamp $t$}
    \STATE Query the corresponding value of the timestamp $t$ to get the imputed data $\hat{\mathbf{X}}=f_{\theta^{*}}(t)$
\ENDFOR
\STATE \textbf{return} $\hat{\mathbf{X}}$
\end{algorithmic}
\label{app_algo_imputeinr}
\end{algorithm}

\begin{table*}[]
\centering
\caption{Details of benchmark datasets.}
\begin{center}
\begin{small}
\scalebox{1}{
\setlength\tabcolsep{3pt}
\begin{tabulary}{0.5cm}{c|c|c|c|c|c|c}
\toprule
Dataset & Source                              & Dimension & Window & \#Training & \#Test & Missing Ratio (\%) \\
\midrule
ETT     & Electricity Transformer Temperature & 7         & 96     & 34465      & 11521 & 0  \\
Weather & Weather Station                     & 21        & 96     & 36696      & 10444 & 0  \\
BAQ     & Beijing Multi-Site Air-Quality      & 132       & 96     & 20448        & 7296 & 1.85     \\
IAQ     & Italy Air Quality                   & 13        & 96     & 5568         & 1824 & 0     \\
Solar   & Solar Alabama                       & 137       & 96     & 26016        & 13248 & 0      \\
Phy2012 & PhysioNet Challenge 2012            & 37        & 48     & 368208       & 115152 & 79.67   \\
Phy2019 & PhysioNet Challenge 2019            & 34        & 48     & 292992       & 91584 & 79.67   \\
MIMIC3 & Medical Center                       & 76        & 48     & 859344       & 155328 & 63.15 \\
\bottomrule
\end{tabulary}}
\end{small}
\end{center}
\label{table_dataset}
\end{table*}

\subsection{Representation Capability of INR for Time Series}
\label{app_motivation}
To evaluate the representation capability of the INR continuous function for time series, we synthetically create a time series dataset and conduct several validation experiments. The synthetic dataset includes four variables, with two variables sampled from a normal distribution with a mean of 0 and a variance of 1, and the other two variables sampled from a normal distribution with a mean of 1 and a variance of 3. In other words, the four variables are generated from two different distributions.

Based on this synthetic dataset, we test the representation capabilities of four different paradigms of INR continuous functions. As shown in Figure \ref{app_fig_motivation}, Model C demonstrates the fastest convergence speed and the best fitting results. This indicates that the representation capability of INR is strongest when both variable clustering and adaptive grouping are used simultaneously. In contrast, Model D has the worst fitting results, suggesting that the correlation information between variables from the same distribution significantly impacts the representation capability of INR. It is worth noting that in our ablation experiments, using variable clustering or adaptive grouping individually also improves the imputation ability. This is because the variable distributions in real datasets are more complex, making it challenging to separate variables belonging to the same cluster into different groups as in the synthetic dataset.

\subsection{Details of Datasets}
Eight time series benchmark datasets are applied to validate the performance of ImputeINR, including Electricity Transformer Temperature (ETT) \cite{zhou2021informer-ett}, Weather Station (Weather) \cite{weather}, Beijing Multi-Site Air-Quality (BAQ) \cite{zhang2017airquality-baq}, Italy Air Quality (IAQ) \cite{vito2016air-iaq}, Solar Alabama (Solar) \cite{solar}, PhysioNet 2012 Challenge (Phy2012) \cite{silva2012physionet-phy2012}, PhysioNet 2019 Challenge (Phy2019) \cite{reyna2019physionet-phy2019}), and MIMIC-III Clinical Database (MIMIC3) \cite{johnson2018mimic}. Table \ref{table_dataset} shows details of the above benchmark datasets. These datasets are collected from different fields and have varying characteristics. Also, these datasets are pre-split into training and test datasets and are publicly available \cite{wutimesnet,du2023pypots}. Based on these datasets, we can evaluate the ability of models to handle varying numbers of variables and different sizes of datasets.
\begin{itemize}
    \item \textbf{ETT} It consists of electricity consumption data, which contains multiple time series from various electricity meters deployed across different regions.
    \item \textbf{Weather} It contains meteorological variables such as temperature, humidity, and wind speed, collected from weather stations.
    \item \textbf{BAQ} It is composed of air quality measurements from different sensors in Beijing, containing data such as particulate matter concentrations (PM10, PM2.5), temperature, and humidity. 
    \item \textbf{IAQ} It contains air quality measurements from various monitoring stations in Italy, including measurements of CO, Non Metanic Hydrocarbons, Benzene, Total Nitrogen Oxides (NOx) and Nitrogen Dioxide (NO2).
    \item \textbf{Solar} It consists of solar power generation data collected from the state of Alabama, which is measured by solar irradiance and power output from photovoltaic panels.
    \item \textbf{Phy2012} It is part of the PhysioNet repository, containing multivariate time series data from ICU patients, which is challenging due to its high missing rate.
    \item \textbf{Phy2019} Similar to Phy2012, it is also from PhysioNet and contains time series data from ICU patients, but with additional variables and a more diverse range of monitoring equipment.
    \item \textbf{MIMIC3} It is a large, freely available database comprising de-identified health-related data associated with over forty thousand patients who stayed in critical care units of the Beth Israel Deaconess Medical Center.
\end{itemize}

\begin{figure*}[]
\centering
\subfloat[Robustness on Mask Rates.]
{
 \centering
 \includegraphics[width=8cm]{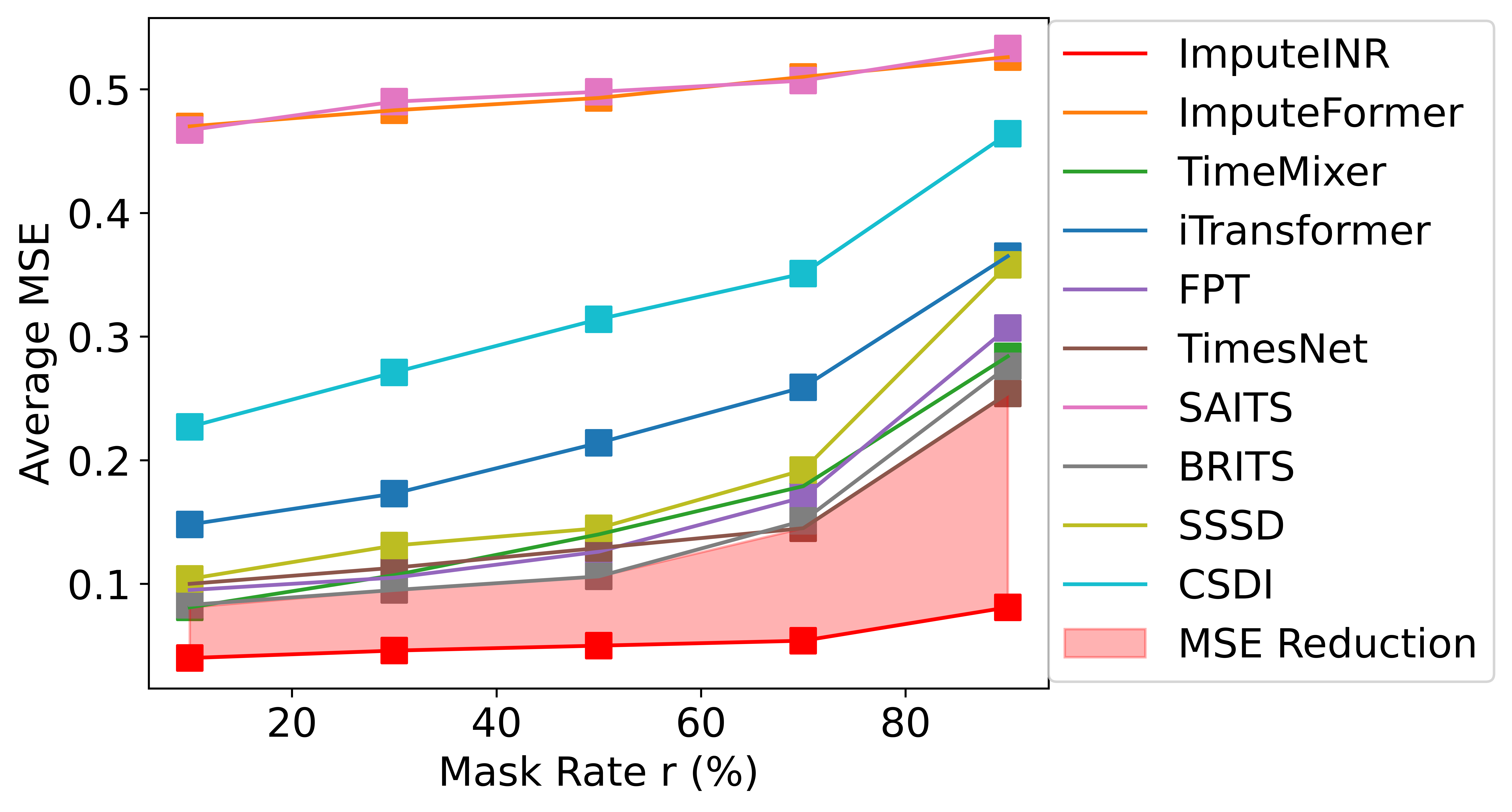}
 \label{fig_robustness_maskrate}
}
\hfill
\subfloat[Robustness on \# of Variables.]
{
 \centering
 \includegraphics[width=8cm]{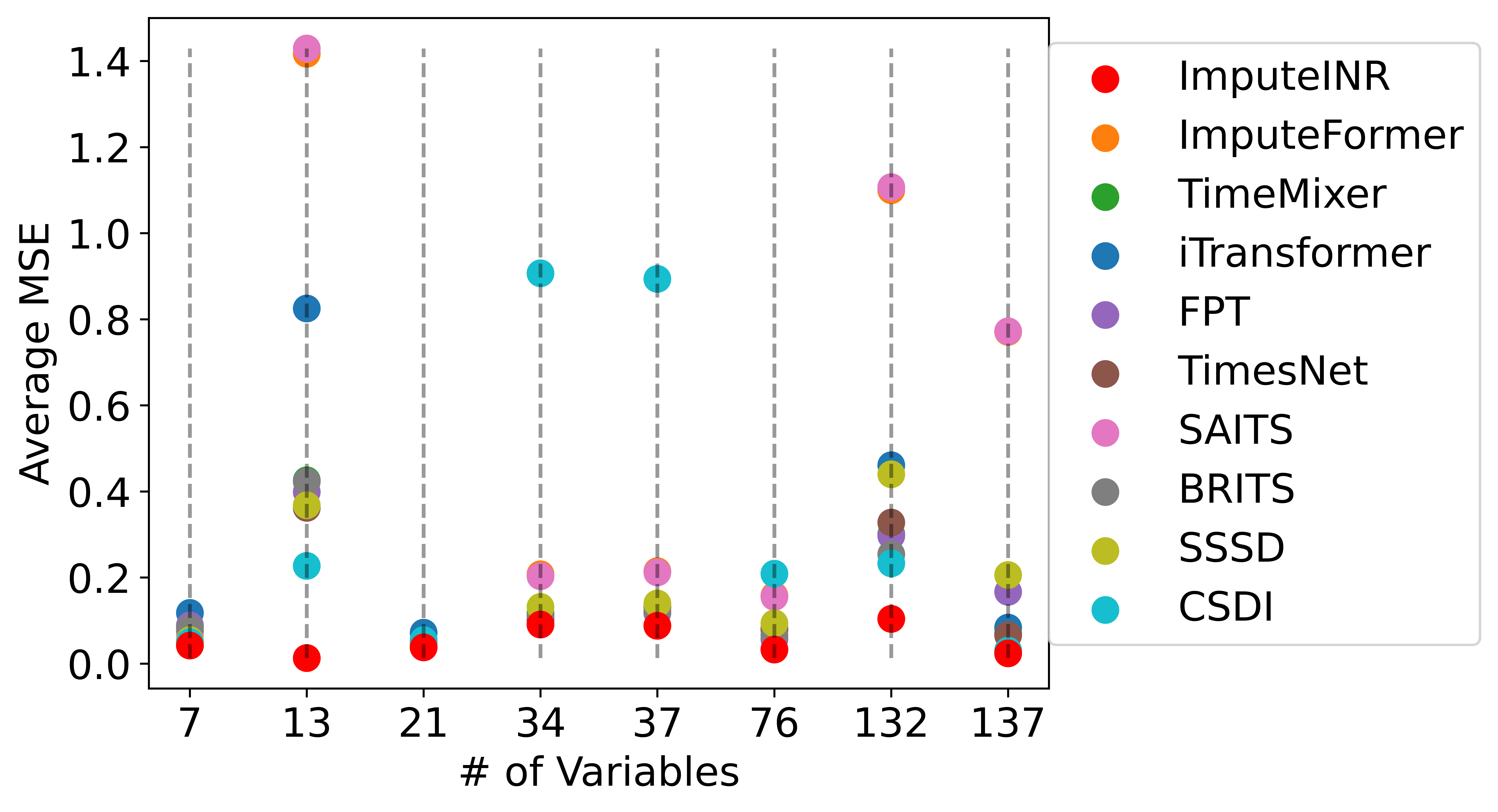}
 \label{fig_robustness_variables}
}
\caption{Robustness analysis for mask rates and the number of variables.}
\label{fig_robustness}
\end{figure*}

\subsection{Details of Baseline Models}
\label{app_baseline}
The details of the baseline models are summarized here.
\begin{itemize}
    \item \textbf{ImputeFormer}\footnote{https://github.com/tongnie/ImputeFormer} A low-rank-induced Transformer that strikes a balance between strong inductive bias and high model expressiveness. By leveraging the inherent structures of spatiotemporal data, ImputeFormer learns well-balanced signal-noise representations, making it adaptable to a wide range of imputation challenges.
    \item \textbf{TimeMixer} \footnote{https://github.com/kwuking/TimeMixer} A fully MLP-based architecture incorporates Past-Decomposable-Mixing and Future-Multipredictor-Mixing blocks to effectively leverage disentangled multiscale series during both past extraction and future prediction phases. 
    \item \textbf{iTransformer} \footnote{https://github.com/thuml/iTransformer} A transformer-based architecture that straightforwardly applies the attention mechanism and feed-forward network to the inverted dimensions. In this approach, the time points of each individual series are embedded into variate tokens, which the attention mechanism uses to capture multivariate correlations. Simultaneously, the feed-forward network operates on each variate token to learn nonlinear representations.
    \item \textbf{FPT} \footnote{https://github.com/DAMO-DI-ML/NeurIPS2023-One-Fits-All/tree/main} A frozen pre-trained transformer that leverages large language models on billions of tokens for time series analysis. Specifically, the self-attention and feedforward layers of the residual blocks in the pre-trained model are retained. It is assessed through fine-tuning across all major types of time series tasks.
    \item \textbf{TimesNet} \footnote{https://github.com/thuml/Time-Series-Library} A method that transforms the 1D time series into a set of 2D tensors based on multiple periods. This transformation can embed the intraperiod- and interperiod-variations into the columns and rows of the 2D tensors respectively, making the 2D-variations to be easily modeled by 2D kernels. 
    \item \textbf{SAITS} \footnote{https://github.com/WenjieDu/SAITS} A self-attention mechanism-based method that learns missing values using a weighted combination of two diagonally-masked self-attention (DMSA) blocks. DMSA effectively captures both temporal dependencies and feature correlations across time steps, enhancing imputation accuracy and training speed. Additionally, the weighted combination allows SAITS to dynamically assign weights to the representations learned from the two DMSA blocks based on the attention map and missingness information.
    \item \textbf{BRITS} \footnote{https://github.com/caow13/BRITS} A RNN-based method directly learns the missing values in a bidirectional recurrent dynamical system, without any specific assumption. 
    \item \textbf{SSSD} \footnote{https://github.com/AI4HealthUOL/SSSD} A diffusion-based method that leverages conditional diffusion models and structured state space models to effectively capture long-term dependencies in time series data.
    \item \textbf{CSDI} \footnote{https://github.com/ermongroup/CSDI} A score-based diffusion model conditioned on observed data allows it to better exploit correlations between observed values.
\end{itemize}

\subsection{Robustness Analysis}
We further evaluate the robustness of our ImputeINR method on mask rate $r$ and the number of variables. As shown in Figure \ref{fig_robustness_maskrate}, ImputeINR outperforms other comparison methods under all mask rate settings, proving its robustness on diverse mask rates. Particularly, as the mask rate $r$ increases, the improvement of our method over others also becomes more pronounced. For example, when $r=10\%$, the average MSE of our method is reduced by 49.5\%, while at $r=0.9$, the reduction reaches 69.2\%. In addition, we also validate the robustness of the number of variables. As shown in Figure \ref{fig_robustness_variables}, our method consistently performs the best on diverse numbers of variables. This demonstrates that our approach effectively addresses the challenges of multi-variable scenarios.

\end{document}